\documentclass[sigconf]{acmart}

\AtBeginDocument{%
  \providecommand\BibTeX{{%
    \normalfont B\kern-0.5em{\scshape i\kern-0.25em b}\kern-0.8em\TeX}}}

\copyrightyear{2026}
\acmYear{2026}
\setcopyright{cc}
\setcctype{by}
\acmConference[WWW '26]{Proceedings of the ACM Web Conference 2026}{April 13--17, 2026}{Dubai, United Arab Emirates}
\acmBooktitle{Proceedings of the ACM Web Conference 2026 (WWW '26), April 13--17, 2026, Dubai, United Arab Emirates}
\acmPrice{}
\acmDOI{10.1145/3774904.3792718}
\acmISBN{979-8-4007-2307-0/2026/04}
\graphicspath{{./images/}}

\usepackage{latexsym}
\usepackage{amsmath}

\usepackage{graphicx}
\usepackage{booktabs}
\usepackage{multirow}
\usepackage{xcolor}
\usepackage{algorithm}
\usepackage{algorithmic}
\usepackage{tcolorbox}
\usepackage{url}

\definecolor{algobg}{RGB}{240, 248, 255}
\definecolor{algoframe}{RGB}{70, 130, 180}
\definecolor{algotitle}{RGB}{25, 25, 112}
\definecolor{commentcolor}{RGB}{34, 139, 34}

\definecolor{algobg2}{RGB}{255, 248, 240}
\definecolor{algoframe2}{RGB}{205, 92, 92}
\definecolor{algotitle2}{RGB}{139, 69, 19}
\definecolor{commentcolor2}{RGB}{178, 34, 34}

\tcbset{
    algorithm/.style={
        enhanced,
        colback=algobg,
        colframe=algoframe,
        coltitle=white,
        colbacktitle=algotitle,
        fonttitle=\bfseries,
        rounded corners=3pt,
        boxrule=1pt,
        drop shadow,
        before skip=10pt,
        after skip=10pt
    },
    algorithm2/.style={
        enhanced,
        colback=algobg2,
        colframe=algoframe2,
        coltitle=white,
        colbacktitle=algotitle2,
        fonttitle=\bfseries,
        rounded corners=3pt,
        boxrule=1pt,
        drop shadow,
        before skip=10pt,
        after skip=10pt
    }
}

\begin{document}


\title[They Said Memes Were Harmless --- We Found the Ones That Hurt: Decoding Jokes, \\ Symbols, and Cultural References]{They Said Memes Were Harmless --- We Found the Ones That Hurt: Decoding Jokes, Symbols, and Cultural References}


\author{Sahil Tripathi}
\email{sahilkrtr@gmail.com}
\affiliation{%
  \institution{Jamia Hamdard}
  \country{New Delhi, India}
}

\author{Gautam Siddharth Kashyap}
\email{gautam.kashyap@hdr.mq.edu.au}
\affiliation{%
  \institution{Macquarie University}
  \country{Sydney, New South Wales, Australia}
}

\author{Mehwish Nasim}
\email{mehwish.nasim@uwa.edu.au}
\affiliation{%
  \institution{The University of Western Australia}
  \country{Perth, Australia}
}

\author{Jian Yang}
\email{jian.yang@mq.edu.au}
\affiliation{%
  \institution{Macquarie University}
  \country{Sydney, New South Wales, Australia}
}

\author{Jiechao Gao}
\authornote{Corresponding Author}
\email{jiechao@stanford.edu}
\affiliation{%
  \institution{Stanford University}
  \country{California, United States}
}

\author{Usman Naseem}
\authornotemark[1]
\email{usman.naseem@mq.edu.au}
\affiliation{%
  \institution{Macquarie University}
  \country{Sydney, New South Wales, Australia}
}

\renewcommand{\shortauthors}{Sahil Tripathi et al.}

\begin{abstract}
Meme-based social abuse detection is challenging because harmful intent often relies on implicit cultural symbolism and subtle cross-modal incongruence. Prior approaches, from fusion-based methods to in-context learning with Large Vision-Language Models (LVLMs), have made progress but remain limited by three factors: i) cultural blindness (missing symbolic context), ii) boundary ambiguity (satire vs.\ abuse confusion), and iii) lack of interpretability (opaque model reasoning). We introduce \texttt{CROSS-ALIGN+}, a three-stage framework that systematically addresses these limitations: (1) \textit{Stage I} mitigates cultural blindness by enriching multimodal representations with structured knowledge from ConceptNet, Wikidata, and Hatebase; (2) \textit{Stage II} reduces boundary ambiguity through parameter-efficient LoRA adapters that sharpen decision boundaries; and (3) \textit{Stage III} enhances interpretability by generating cascaded explanations. Extensive experiments on five benchmarks and eight LVLMs demonstrate that \texttt{CROSS-ALIGN+} consistently outperforms state-of-the-art methods, achieving up to 17\% relative F1 improvement while providing interpretable justifications for each decision.
\end{abstract}




\maketitle

\begin{figure}[t!]
    \centering
    \includegraphics[width=5cm]{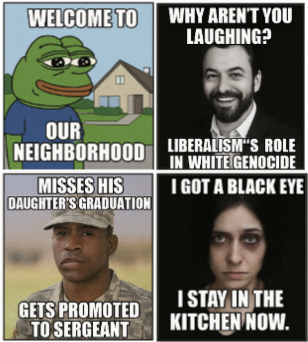}
    \vspace{-0.3cm}
    \caption{Toy Example of Implicit Abuse in Memes. 
Illustrative samples showing how memes encode harmful intent through subtle cross-modal incongruence and cultural symbolism. Although humorous in isolation, cultural context exposes exclusionary or abusive meanings—e.g., the alt-right appropriation of ``Pepe the Frog'', gendered subservience, or racially charged imagery—highlighting the need for culturally grounded multimodal reasoning in \texttt{CROSS-ALIGN+}.
}
    \label{Toy Example}
\end{figure}

\section{Introduction}
\label{sec:introduction}

Memes have become one of the most viral forms of communication on the web, shaping online discourse on a wide range of platforms including X (Twitter), Reddit, and TikTok~\cite{shah2024memeclip,pramanick2021detecting}. Although often humorous, memes are increasingly weaponized to propagate hate, exclusion, and disinformation (see Figure \ref{Toy Example})~\cite{kiela2020hateful,pramanick2021detecting}, exploiting their ability to encode harmful intent through subtle cross-modal cues. This makes meme-based abuse detection a critical challenge for web-scale content moderation systems~\cite{lin2024goat,lin2025ask}. Detecting abusive memes requires reasoning about \textit{implicit cultural symbolism} and \textit{cross-modal incongruence}~\cite{kiela2020hateful,pramanick2021detecting}. For example, a meme featuring Pepe the Frog with the caption ``Welcome to our neighborhood'' appears harmless in isolation, but cultural context reveals Pepe's co-optation as an alt-right symbol, transforming the message into exclusionary propaganda~\cite{lynch2022memes}. Current systems fail because they cannot ground such visual-textual signals within broader cultural knowledge~\cite{xu2024lemma}.

Recent advances in Large Vision-Language Models (LVLMs)~\cite{liu2023visual,bai2023qwen,liu2023llava} offer strong multimodal understanding, yet remain limited in meme abuse detection due to: (i) cultural blindness \textemdash missing symbolic context and cultural references~\cite{sharma2022disarm,shah2024memeclip}; (ii) boundary ambiguity \textemdash confusing satire with abuse~\cite{yang2023invariant,aggarwal2024text}; and (iii) lack of interpretability \textemdash providing opaque predictions without explanations~\cite{zhang2024critic,balazevic2023towards}. Previous approaches from fusion-based methods (e.g. DISARM~\cite{sharma2022disarm}, MemeCLIP~\cite{shah2024memeclip}) to in-context learning strategies~\cite{balazevic2023towards,zhang2024critic}, have improved performance, but do not address these limitations systematically~\cite{xuan2024lemma,lin2025ask}.

\textbf{Our Approach.} We propose \texttt{CROSS-ALIGN+} (\textit{Cross-modal Social Signal Alignment with External Knowledge \& Adaptive Contrastive Reasoning}), a three-stage framework that directly mitigates the core limitations of existing LVLM-based methods. \textbf{Stage I}(\textit{External Knowledge Grounding}), enriches multimodal representations with structured cultural knowledge from ConceptNet~\cite{speer2017conceptnet}, Wikidata~\cite{vrandevcic2014wikidata}, and Hatebase~\cite{hatebase2024}, thereby addressing cultural blindness. \textbf{Stage II}(\textit{Adaptive Contrastive Fine-Tuning}), leverages parameter-efficient LoRA adapters~\cite{hu2021lora} with contrastive objectives to sharpen decision boundaries and reduce boundary ambiguity between satire and abuse. \textbf{Stage III}(\textit{Hierarchical Multistage Reasoning}), introduces cascaded natural language explanations that enhance interpretability and trustworthiness~\cite{wei2022chain,zhang2024critic}. In summary, this work makes three main contributions:
\begin{itemize}
    \item We present the first framework that systematically integrates external cultural knowledge into LVLM-based meme abuse detection. 
    \item We design a three-stage framework that explicitly aligns each stage with a core limitation: cultural blindness (Stage I), boundary ambiguity (Stage II), and lack of interpretability (Stage III).  
    \item We conduct extensive empirical analysis across five benchmarks and eight LVLMs, showing up to 17\% relative F1 improvements and strong gains on culturally grounded abuse patterns. 
\end{itemize}

\section{Related Works}
\label{sec:related}

Our work is based on two key lines of research: i) detection of multimodal abuse with LVLMs, and ii) approaches to knowledge integration and efficient model adaptation. We position \texttt{CROSS-ALIGN+} within these areas to highlight the unique gaps it addresses.  

\subsection{Multimodal Abuse Detection and LVLMs}  

Early multimodal abuse detection methods relied on feature fusion of visual and textual signals~\cite{kiela2020hateful}, but such approaches struggle with implicit abuse where harmful intent emerges from subtle cross-modal incongruence~\cite{aggarwal2024text}. More recent models leverage pre-trained LVLMs, such as MemeCLIP~\cite{shah2024memeclip}, and specialized designs like DISARM~\cite{sharma2022disarm} for victim-aware detection or invariant learning for robustness~\cite{yang2023invariant}. While these advances improve performance, they lack the ability to ground interpretations in cultural context, a key factor for understanding symbolic abuse (e.g., political or historical symbols).  In-context learning has been explored as a lightweight way to adapt LVLMs without parameter updates~\cite{wei2022chain}. Extensions such as Hummingbird~\cite{balazevic2023towards} and Critic-V~\cite{zhang2024critic} show promise for retrieval-augmented prompting and error detection. However, these approaches remain constrained by semantic similarity, often missing cases where abusive meaning depends on culturally loaded symbolism. \texttt{CROSS-ALIGN+} directly addresses this gap by systematically linking LVLM representations to structured cultural knowledge and aligning decisions with symbolic reasoning.  

\subsection{Knowledge Integration and Efficient Adaptation}  

Knowledge integration has proven to be effective in NLP tasks such as factual reasoning~\cite{petroni2019language}. For multimodal understanding, LEMMA~\cite{xuan2024lemma} recently demonstrated that external knowledge can enhance misinformation detection. Yet, systematic integration of cultural knowledge for abuse detection remains underexplored. Resources such as ConceptNet~\cite{speer2017conceptnet}, Wikidata~\cite{vrandevcic2014wikidata}, and Hatebase~\cite{hatebase2024} provide complementary symbolic knowledge, but prior works have not combined them in a principled pipeline for meme abuse detection. \texttt{CROSS-ALIGN+} fills this gap by using these resources to contextualize visual-textual signals with cultural grounding.  In parallel, parameter-efficient fine-tuning methods such as adapters~\cite{houlsby2019parameter} and LoRA~\cite{hu2021lora} enable model adaptation without prohibitive compute. These techniques have been adopted for multimodal tasks but have rarely been explored in the context of contrastive alignment for abuse detection. \texttt{CROSS-ALIGN+} extends this line by coupling LoRA-based contrastive fine-tuning with knowledge-grounded demonstrations, improving decision boundaries while preserving LVLM generalization.  

\begin{figure*}[t]
\vspace{-0.4cm}
    \centering
    \includegraphics[width=14cm, height=8cm]{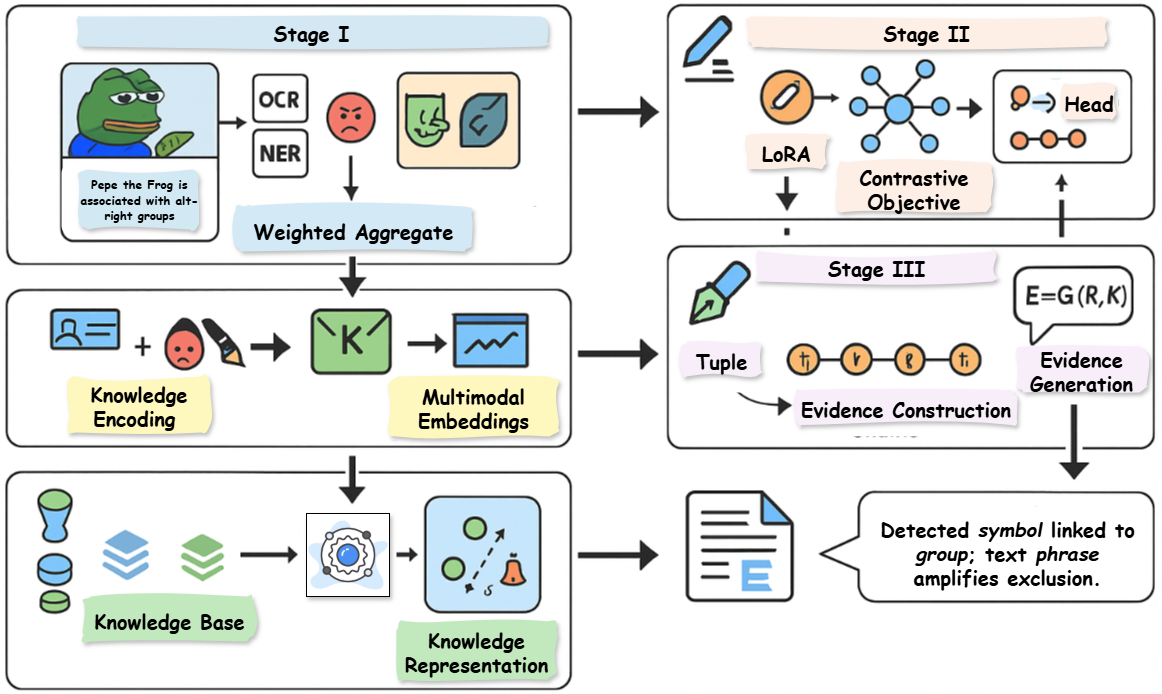}
    \vspace{-0.4cm}
    \caption{\textbf{Overview of CROSS-ALIGN+.} The pipeline consists of three stages: (I) external knowledge grounding using ConceptNet, Wikidata, and Hatebase to enrich multimodal representations; (II) adaptive contrastive fine-tuning with LoRA to improve boundary discrimination; and (III) hierarchical multistage reasoning to generate interpretable explanations. CROSS-ALIGN+ outputs both abuse predictions and faithful justifications for transparent meme moderation.}
    \label{fig:framework}
\end{figure*}

\section{Methodology}
\subsection{Problem Formulation}
\label{sec:problem}

We consider meme-based social abuse detection as a binary classification problem augmented with external cultural knowledge. Each meme is a pair $(I, T)$ with an image $I \in \mathbb{R}^{H \times W \times 3}$ and an overlaid/associated text $T$. The goal is to predict $\hat{y} \in \{0,1\}$ indicating \textit{abusive} $(1)$ vs.\ \textit{benign} $(0)$ content as shown in Equation (1), where $\Theta$ are model parameters and $\mathcal{K}$ denotes external cultural knowledge.
\begin{equation}
\hat{y} = f_\Theta(I, T, \mathcal{K}); \qquad
f_\Theta: (\mathbb{R}^{H \times W \times 3}, \mathcal{V}^*) \times \mathcal{K} \rightarrow \{0,1\},
\end{equation}

\noindent\textbf{Why knowledge is required.}
Abusive meaning often depends on \emph{cross-modal incongruence} and \emph{culturally loaded symbols}. Thus, $f_\Theta$ should depend not only on $(I,T)$ but also on knowledge items $K \subset \mathcal{K}$ that contextualize entities in $(I,T)$ as shown in Equation (2), where $\mathcal{E}(I,T)$ extracts entities from both modalities, $\phi$ links entities to KB entries, and $\mathrm{Retrieve}$ gathers culturally relevant facts/symbols.
\begin{equation}
\hat{y} = f_\Theta\big(I, T, K(I,T)\big), \quad
K(I,T) = \mathrm{Retrieve}\big(\phi(\mathcal{E}(I,T))\big),
\end{equation}

\noindent\textbf{Learning.}
Given a training set $\mathcal{D}=\{(I_i, T_i, y_i)\}_{i=1}^N$, we learn $\Theta$ by minimizing a classification objective, and—crucially—aligning representations with a contrastive objective that separates abusive/benign cases under knowledge context as shown in Equation (3), where $\hat{p}_i = \sigma(g_\Theta(I_i,T_i,K_i))$ are model probabilities and $\lambda > 0$ balances the losses. Section~\ref{sec:method} details the construction of $K_i$ and the contrastive term. 
\begin{equation}
\small
\mathcal{L} = \mathcal{L}_{\mathrm{cls}} + \lambda\, \mathcal{L}_{\mathrm{contrast}},
\qquad
\mathcal{L}_{\mathrm{cls}} = -\frac{1}{N}\sum_{i=1}^N \big[y_i \log \hat{p}_i + (1-y_i)\log (1-\hat{p}_i)\big],
\end{equation}


\subsection{\texttt{CROSS-ALIGN+}}
\label{sec:method}

\texttt{CROSS-ALIGN+} (see Figure \ref{fig:framework}) is a three-stage framework aligned \emph{one-to-one} with the limitations identified in Section~\ref{sec:introduction}: (I) cultural blindness, (II) boundary ambiguity, and (III) lack of interpretability. Each stage receives the output from the previous one, transforms that with stage-specific objectives, and passes enriched representations forward. Let $\mathrm{LVLM}_\theta$ denote a pre-trained LVLM encoder producing a multimodal representation $\mathbf{h}_{\mathrm{mm}}$ from $(I,T)$.  

\subsubsection{Stage I: External Knowledge Grounding}
\label{subsec:stage1}

Stage I mitigates \textit{cultural blindness} by grounding multimodal inputs in structured cultural knowledge. Its output is a knowledge-enriched representation $\tilde{\mathbf{h}}$ that is passed to Stage II.  

\textbf{Entity Extraction and Linking.} We extract entities from both modalities: $
\mathcal{E}_I = \mathrm{OCR/OD}(I), \quad
\mathcal{E}_T = \mathrm{NER}(T), \quad
\mathcal{E} = \mathcal{E}_I \cup \mathcal{E}_T$. Entities are linked via $\phi: \mathcal{E} \rightarrow \mathcal{K}$ to entries in ConceptNet~\cite{speer2017conceptnet}, Wikidata~\cite{vrandevcic2014wikidata}, and Hatebase~\cite{hatebase2024}. For each $e \in \mathcal{E}$ we retrieve source-specific knowledge $\{k_{e,c}, k_{e,w}, k_{e,h}\}$ and construct a weighted aggregate: $
K = \sum_{e \in \mathcal{E}} \big( w_c\, k_{e,c} + w_w\, k_{e,w} + w_h\, k_{e,h} \big)$.

\textbf{Knowledge Encoding and Fusion.} We encode $K$ using the text encoder native to the chosen LVLM (e.g., if DeepSeek-VL-S is used, its own text encoder performs the encoding), which processes linearized knowledge triples (e.g., ``Pepe the Frog is associated with alt-right groups’’) into embeddings. This yields $\mathbf{M}_K \in \mathbb{R}^{L_K \times d}$ aligned with the multimodal representation $\mathbf{h}_{\mathrm{mm}}$. Given $\mathbf{h}_{\mathrm{mm}} \in \mathbb{R}^{L \times d}$ from $\mathrm{LVLM}_\theta(I,T)$, we fuse knowledge using cross-attention as shown in Equation (4), where $\tilde{\mathbf{h}}$ is the knowledge-grounded representation. A lightweight \emph{gate} can optionally modulate the residual: $
\tilde{\mathbf{h}} = \mathbf{h}_{\mathrm{mm}} + \sigma\!\left(\mathbf{W}_g[\mathbf{h}_{\mathrm{mm}};\mathrm{pool}(\mathbf{M}_K)]\right) \odot \mathrm{CrossAttn}(\cdot)
$. This $\tilde{\mathbf{h}}$ is the input to Stage II.
\begin{equation}
\tilde{\mathbf{h}}
= \mathbf{h}_{\mathrm{mm}} + \mathrm{CrossAttn}\big(Q=\mathbf{h}_{\mathrm{mm}}, K=\mathbf{M}_K, V=\mathbf{M}_K\big),
\label{eq:knowledge-fusion}
\end{equation}  

\subsubsection{Stage II: Adaptive Contrastive Fine-Tuning}
\label{subsec:stage2}

Stage II mitigates \textit{boundary ambiguity} by refining $\tilde{\mathbf{h}}$ with parameter-efficient contrastive adaptation. It produces a refined representation and prediction probability $\hat{p}$, both of which were used in Stage III.  

\textbf{LoRA Integration.}
We adapt the attention projections in $\mathrm{LVLM}_\theta$ with LoRA~\cite{hu2021lora}. For an attention weight $\mathbf{W} \in \mathbb{R}^{d \times d}$, $
\mathbf{W}' = \mathbf{W} + \alpha\, \mathbf{B}\mathbf{A}, \quad \mathbf{B} \in \mathbb{R}^{d \times r},\ \mathbf{A} \in \mathbb{R}^{r \times d},\ r \ll d$, where $\alpha$ scales the low-rank update. LoRA is applied to self-and cross-attention layers operating on $\tilde{\mathbf{h}}$ and $\mathbf{M}_K$.  

\textbf{Contrastive Objective.}
Let $\mathbf{z}_i = \mathrm{Pool}(\tilde{\mathbf{h}}_i)$ be the pooled embedding for sample $i$. To separate abusive from benign memes, we apply a triplet loss with cosine distance $d(\cdot,\cdot)$ as shown in Equation (5), where $\mathbf{z}_i^+$ is a positive (same class, culturally similar) and $\mathbf{z}_i^-$ a negative (different class, culturally close), with margin $\delta>0$. Triplets are chosen via a hybrid similarity see Equation (6).
\begin{equation}
\mathcal{L}_{\mathrm{contrast}}
= \frac{1}{|\mathcal{B}|}\sum_{i \in \mathcal{B}}
\max\big(0,\ \delta + d(\mathbf{z}_i,\mathbf{z}_i^+) - d(\mathbf{z}_i,\mathbf{z}_i^-)\big),
\label{eq:triplet}
\end{equation}
\begin{equation}
s(i,j) = \lambda_s\, \mathrm{sim}(\mathbf{h}_{\mathrm{mm}}^{(i)}, \mathbf{h}_{\mathrm{mm}}^{(j)}) +
\lambda_c\, \mathrm{cultRel}(\mathcal{E}^{(i)}, \mathcal{E}^{(j)}),\quad \lambda_s+\lambda_c=1.
\end{equation}

\textbf{Classification Head.}
A lightweight head (i.e. sigmoid) $g_\Theta(\cdot)$ predicts $\hat{p}=\sigma(g_\Theta(\tilde{\mathbf{h}}))$, with cross-entropy loss $\mathcal{L}_{\mathrm{cls}}$. The total loss is: $\mathcal{L} = \mathcal{L}_{\mathrm{cls}} + \lambda\, \mathcal{L}_{\mathrm{contrast}}$. The refined representation $\tilde{\mathbf{h}}$ and prediction $\hat{p}$ form the input to Stage III.  

\subsubsection{Stage III: Hierarchical Multistage Reasoning}
\label{subsec:stage3}

Stage III mitigates the \textit{lack of interpretability} by generating explanations conditioned on the refined representation and knowledge. Its output is a natural-language justification $E$.  

\textbf{Evidence Construction.}
From linked knowledge $K$ and attention weights in Eq.~\ref{eq:knowledge-fusion}, we extract evidence tuples $(e,\ \mathrm{symbol}(e)$, relation, snippet). Salience is derived from cross-attention and entity spans in $(I,T)$.  

\textbf{Reasoning Chains.}
We select top-$m$ evidence steps $\mathcal{R}=\{r_1,r_m\}$ with scores: $
w_i = \mathrm{softmax}\!\left(\mathbf{W}_r \,[\,\mathrm{pool}(\tilde{\mathbf{h}});\ \mathbf{e}_i;\ \mathbf{k}_i\,]\right)$.

\textbf{Explanation Generation.}
Reasoning chains are realized into explanations using constrained templates: $E = G(\mathcal{R}, \mathcal{E}, K)$, ensuring faithfulness (e.g., ``Detected \emph{symbol} linked to \emph{group}; text \emph{phrase} amplifies exclusion.'') (see Section \ref{subsec:explanation-analysis}).  

\section{Experimental Setup}
\label{sec:experiments}

We evaluate \texttt{CROSS-ALIGN+} across five benchmark datasets, eight state-of-the-art LVLMs, and multiple baselines. Training proceeds in a parameter-efficient manner: for each mini-batch, entities are extracted and linked to external knowledge bases; the retrieved cultural knowledge is fused with multimodal representations (Stage I); the representation is refined using LoRA adapters (Stage II) to produce predictions; and triplets are constructed to apply the contrastive loss, with updates restricted to LoRA modules and a lightweight classification head. At inference time, unseen memes follow a parallel flow—knowledge retrieval and fusion yield a grounded representation, which is refined to generate predictions and cascaded explanations (Stage III). To ensure scalability, knowledge lookups are cached and reused across examples. Despite the three-stage design, computational overhead remains modest: LoRA adds less than 1\% additional parameters, and knowledge fusion scales as $O(L \cdot L_K)$ with $L_K$ bounded by entity-driven retrieval (see Section \ref{subsec:efficiency}).

\subsection{Datasets}
\label{subsec:datasets}

We evaluate \texttt{CROSS-ALIGN+} on \textbf{GOAT-Bench}~\cite{lin2024goat}, a recently released multimodal benchmark that consolidates several widely used datasets for harmful meme detection. Each dataset represents a distinct form of abusive online expression, allowing us to test generalization across heterogeneous contexts. For consistency with GOAT-Bench reporting, we present results under five task categories: \textit{Harmfulness}, \textit{Hatefulness}, \textit{Misogyny}, \textit{Offensiveness}, and \textit{Sarcasm}. These categories directly correspond to the datasets described below, i.e., Harm-C/P $\rightarrow$ Harmfulness, FHM $\rightarrow$ Hatefulness, MAMI $\rightarrow$ Misogyny, MultiOFF $\rightarrow$ Offensiveness, and MSD $\rightarrow$ Sarcasm. 

\begin{itemize}

    \item \textbf{Harm-C/P (Harmful Content / Propaganda)}~\cite{lin2024goat}: 1,063 memes labeled for socially harmful narratives such as propaganda, exclusionary discourse, and systemic disinformation. Unlike the other datasets that focus on interpersonal abuse, Harm-C/P targets collective or societal harms, highlighting the need for cultural knowledge integration.
    
    \item \textbf{FHM (Hateful Memes)}~\cite{kiela2020hateful}: 2,000 memes where hate is often expressed through subtle image–text incongruence. Each modality may appear benign in isolation, but their combination encodes harmful stereotypes or exclusionary meaning. This dataset stresses cross-modal reasoning.  

    \item \textbf{MAMI (Multimodal Misogyny)}~\cite{fersini2022mami}: 1,000 memes annotated for sexist and gender-based abuse. It includes both explicit misogynistic content and implicit cases such as irony, stereotypes, or objectification. This dataset requires detecting culturally nuanced and symbolic signals of bias.  

    \item \textbf{MultiOFF (Multimodal Offensiveness)}~\cite{das2020multi}: 743 memes labeled for offensive content. Unlike FHM and MAMI, MultiOFF captures broader offensive language and imagery not tied to a single protected group. It evaluates whether models can generalize to toxicity expressed in more colloquial or coded forms.  

    \item \textbf{MSD (Multimodal Sarcasm Detection)}~\cite{cai2019sarcasm}: 1,820 memes annotated for sarcasm. Sarcasm blurs the line between benign humor and abuse, often requiring pragmatic and cultural knowledge to distinguish playful satire from harmful commentary. This dataset emphasizes boundary ambiguity.    
\end{itemize}

\subsection{Evaluation Metrics}
\label{subsec:metrics}

We adopt two standard metrics for binary abuse detection. All reported values are expressed as percentages (\%). \textbf{Accuracy (Acc)} quantifies the overall proportion of correctly classified memes. \textbf{Macro-F1 (F1)} balances precision and recall across both classes by averaging their individual F1 scores. Macro-F1 is particularly important in abuse detection, as abusive samples are often underrepresented and accuracy alone may be biased toward the majority class.  

\begin{table*}[t]
\caption{Performance of 8 LVLMs under Vanilla Zero-Shot and Standard ICL, with and without \texttt{CROSS-ALIGN+}. Superscripts indicate absolute gain (+) or drop (–) compared to the baseline. Overall score in Table is the macro-average across these five categories.}
\vspace{-0.4cm}
\label{tab:main_results}
\centering
\footnotesize
\resizebox{\textwidth}{!}{
\begin{tabular}{l|cc|cc|cc|cc|cc|cc}
\hline
\multirow{2}{*}{Models} &
\multicolumn{2}{c|}{Harmfulness} &
\multicolumn{2}{c|}{Hatefulness} &
\multicolumn{2}{c|}{Misogyny} &
\multicolumn{2}{c|}{Offensiveness} &
\multicolumn{2}{c|}{Sarcasm} &
\multicolumn{2}{c}{Overall} \\
 & Acc. (\%) & F1 (\%) & Acc. (\%) & F1 (\%) & Acc. (\%) & F1 (\%) & Acc. (\%) & F1 (\%) & Acc. (\%) & F1 (\%) & Acc. (\%) & F1 (\%) \\
\hline
\multicolumn{13}{c}{\textbf{Vanilla Zero-Shot Inference}} \\
\hline
LLaVA-V1.6 (7B) & 56.9 & 54.7 & 63.1 & 60.5 & 68.8 & 66.2 & 57.0 & 53.8 & 59.2 & 55.6 & 61.0 & 58.2 \\
\quad + CROSS-ALIGN+ & \textbf{67.5}\textcolor{green!80!black}{$^{+10.6}$} & \textbf{65.3}\textcolor{green!80!black}{$^{+10.6}$} & \textbf{74.2}\textcolor{green!80!black}{$^{+11.1}$} & \textbf{71.4}\textcolor{green!80!black}{$^{+10.9}$} & \textbf{80.1}\textcolor{green!80!black}{$^{+11.3}$} & \textbf{78.5}\textcolor{green!80!black}{$^{+12.3}$} & \textbf{69.3}\textcolor{green!80!black}{$^{+12.3}$} & \textbf{66.2}\textcolor{green!80!black}{$^{+12.4}$} & \textbf{71.4}\textcolor{green!80!black}{$^{+12.2}$} & \textbf{69.2}\textcolor{green!80!black}{$^{+13.6}$} & \textbf{72.5}\textcolor{green!80!black}{$^{+11.5}$} & \textbf{70.1}\textcolor{green!80!black}{$^{+11.9}$} \\
Qwen2.5-VL (7B) & 57.8 & 55.2 & 64.5 & 61.0 & 70.2 & 68.0 & 58.1 & 54.4 & 60.0 & 56.7 & 62.1 & 59.0 \\
\quad + CROSS-ALIGN+ & \textbf{69.5}\textcolor{green!80!black}{$^{+11.7}$} & \textbf{67.1}\textcolor{green!80!black}{$^{+11.9}$} & \textbf{77.0}\textcolor{green!80!black}{$^{+12.5}$} & \textbf{74.3}\textcolor{green!80!black}{$^{+13.3}$} & \textbf{83.4}\textcolor{green!80!black}{$^{+13.2}$} & \textbf{81.8}\textcolor{green!80!black}{$^{+13.8}$} & \textbf{71.0}\textcolor{green!80!black}{$^{+12.9}$} & \textbf{67.8}\textcolor{green!80!black}{$^{+13.4}$} & \textbf{73.1}\textcolor{green!80!black}{$^{+13.1}$} & \textbf{70.5}\textcolor{green!80!black}{$^{+13.8}$} & \textbf{74.8}\textcolor{green!80!black}{$^{+12.7}$} & \textbf{72.3}\textcolor{green!80!black}{$^{+13.3}$} \\
InternVL2 (7B) & 55.9 & 53.1 & 61.2 & 57.9 & 67.1 & 64.4 & 54.7 & 51.2 & 57.5 & 54.8 & 59.3 & 56.3 \\
\quad + CROSS-ALIGN+ & \textbf{67.0}\textcolor{green!80!black}{$^{+11.1}$} & \textbf{64.6}\textcolor{green!80!black}{$^{+11.5}$} & \textbf{73.5}\textcolor{green!80!black}{$^{+12.3}$} & \textbf{70.8}\textcolor{green!80!black}{$^{+12.9}$} & \textbf{80.2}\textcolor{green!80!black}{$^{+13.1}$} & \textbf{78.6}\textcolor{green!80!black}{$^{+14.2}$} & \textbf{68.9}\textcolor{green!80!black}{$^{+14.2}$} & \textbf{65.7}\textcolor{green!80!black}{$^{+14.5}$} & \textbf{70.0}\textcolor{green!80!black}{$^{+12.5}$} & \textbf{67.6}\textcolor{green!80!black}{$^{+12.8}$} & \textbf{71.9}\textcolor{green!80!black}{$^{+12.6}$} & \textbf{69.3}\textcolor{green!80!black}{$^{+13.0}$} \\
Gemma-VL (7B) & 52.6 & 49.8 & 58.7 & 54.9 & 64.9 & 62.1 & 51.4 & 48.6 & 54.3 & 50.9 & 56.4 & 53.3 \\
\quad + CROSS-ALIGN+ & \textbf{64.3}\textcolor{green!80!black}{$^{+11.7}$} & \textbf{61.6}\textcolor{green!80!black}{$^{+11.8}$} & \textbf{71.4}\textcolor{green!80!black}{$^{+12.7}$} & \textbf{68.3}\textcolor{green!80!black}{$^{+13.4}$} & \textbf{78.5}\textcolor{green!80!black}{$^{+13.6}$} & \textbf{76.8}\textcolor{green!80!black}{$^{+14.7}$} & \textbf{66.0}\textcolor{green!80!black}{$^{+14.6}$} & \textbf{62.8}\textcolor{green!80!black}{$^{+14.2}$} & \textbf{68.9}\textcolor{green!80!black}{$^{+14.6}$} & \textbf{65.9}\textcolor{green!80!black}{$^{+15.0}$} & \textbf{69.8}\textcolor{green!80!black}{$^{+13.4}$} & \textbf{67.3}\textcolor{green!80!black}{$^{+14.0}$} \\
DeepSeek-VL-S (7B) & 54.9 & 51.8 & 60.3 & 56.7 & 66.2 & 63.5 & 53.2 & 49.9 & 56.7 & 53.6 & 58.7 & 55.1 \\
\quad + CROSS-ALIGN+ & \textbf{66.8}\textcolor{green!80!black}{$^{+11.9}$} & \textbf{63.5}\textcolor{green!80!black}{$^{+11.7}$} & \textbf{72.9}\textcolor{green!80!black}{$^{+12.6}$} & \textbf{70.1}\textcolor{green!80!black}{$^{+13.4}$} & \textbf{79.0}\textcolor{green!80!black}{$^{+12.8}$} & \textbf{77.3}\textcolor{green!80!black}{$^{+13.8}$} & \textbf{67.6}\textcolor{green!80!black}{$^{+14.4}$} & \textbf{64.3}\textcolor{green!80!black}{$^{+14.4}$} & \textbf{69.5}\textcolor{green!80!black}{$^{+12.8}$} & \textbf{66.7}\textcolor{green!80!black}{$^{+13.1}$} & \textbf{70.6}\textcolor{green!80!black}{$^{+13.2}$} & \textbf{67.8}\textcolor{green!80!black}{$^{+12.7}$} \\
MiniGPT-4 (7B) & 53.8 & 51.0 & 59.6 & 56.3 & 65.4 & 62.9 & 52.4 & 49.3 & 55.7 & 52.4 & 57.4 & 54.4 \\
\quad + CROSS-ALIGN+ & \textbf{65.2}\textcolor{green!80!black}{$^{+11.4}$} & \textbf{62.6}\textcolor{green!80!black}{$^{+11.6}$} & \textbf{71.8}\textcolor{green!80!black}{$^{+12.2}$} & \textbf{69.1}\textcolor{green!80!black}{$^{+12.8}$} & \textbf{77.5}\textcolor{green!80!black}{$^{+12.1}$} & \textbf{75.9}\textcolor{green!80!black}{$^{+13.0}$} & \textbf{66.5}\textcolor{green!80!black}{$^{+14.1}$} & \textbf{63.0}\textcolor{green!80!black}{$^{+13.7}$} & \textbf{68.1}\textcolor{green!80!black}{$^{+12.4}$} & \textbf{65.5}\textcolor{green!80!black}{$^{+13.1}$} & \textbf{69.8}\textcolor{green!80!black}{$^{+12.4}$} & \textbf{67.0}\textcolor{green!80!black}{$^{+12.6}$} \\
mPLUG-Owl2-Small (7B) & 55.6 & 52.5 & 61.5 & 58.1 & 67.2 & 64.3 & 54.1 & 51.0 & 57.0 & 53.9 & 59.1 & 56.0 \\
\quad + CROSS-ALIGN+ & \textbf{66.9}\textcolor{green!80!black}{$^{+11.3}$} & \textbf{64.0}\textcolor{green!80!black}{$^{+11.5}$} & \textbf{73.6}\textcolor{green!80!black}{$^{+12.1}$} & \textbf{70.9}\textcolor{green!80!black}{$^{+12.8}$} & \textbf{79.6}\textcolor{green!80!black}{$^{+12.4}$} & \textbf{77.9}\textcolor{green!80!black}{$^{+13.6}$} & \textbf{68.2}\textcolor{green!80!black}{$^{+14.1}$} & \textbf{65.0}\textcolor{green!80!black}{$^{+14.0}$} & \textbf{70.1}\textcolor{green!80!black}{$^{+13.1}$} & \textbf{67.4}\textcolor{green!80!black}{$^{+13.5}$} & \textbf{71.7}\textcolor{green!80!black}{$^{+12.6}$} & \textbf{69.0}\textcolor{green!80!black}{$^{+13.0}$} \\
BLIP-2 OPT (6.7B) & 52.3 & 49.7 & 57.9 & 54.3 & 63.8 & 61.4 & 50.7 & 47.9 & 54.0 & 50.7 & 55.7 & 52.8 \\
\quad + CROSS-ALIGN+ & \textbf{63.9}\textcolor{green!80!black}{$^{+11.6}$} & \textbf{61.2}\textcolor{green!80!black}{$^{+11.5}$} & \textbf{70.5}\textcolor{green!80!black}{$^{+12.6}$} & \textbf{67.4}\textcolor{green!80!black}{$^{+13.1}$} & \textbf{76.1}\textcolor{green!80!black}{$^{+12.3}$} & \textbf{74.3}\textcolor{green!80!black}{$^{+12.9}$} & \textbf{65.0}\textcolor{green!80!black}{$^{+14.3}$} & \textbf{61.9}\textcolor{green!80!black}{$^{+14.0}$} & \textbf{67.3}\textcolor{green!80!black}{$^{+13.3}$} & \textbf{64.5}\textcolor{green!80!black}{$^{+13.8}$} & \textbf{68.5}\textcolor{green!80!black}{$^{+12.8}$} & \textbf{65.9}\textcolor{green!80!black}{$^{+13.1}$} \\
\hline
\multicolumn{13}{c}{\textbf{Standard In-Context Learning (ICL)}} \\
\hline
LLaVA-V1.6 (7B) & 61.7 & 59.3 & 68.4 & 64.9 & 73.2 & 71.8 & 61.1 & 57.7 & 63.8 & 61.4 & 65.6 & 63.0 \\
\quad + CROSS-ALIGN+ & \textbf{72.8}\textcolor{green!80!black}{$^{+11.1}$} & \textbf{70.5}\textcolor{green!80!black}{$^{+11.2}$} & \textbf{78.9}\textcolor{green!80!black}{$^{+10.5}$} & \textbf{76.3}\textcolor{green!80!black}{$^{+11.4}$} & \textbf{84.0}\textcolor{green!80!black}{$^{+10.8}$} & \textbf{82.6}\textcolor{green!80!black}{$^{+10.8}$} & \textbf{72.0}\textcolor{green!80!black}{$^{+10.9}$} & \textbf{69.1}\textcolor{green!80!black}{$^{+11.4}$} & \textbf{74.9}\textcolor{green!80!black}{$^{+11.1}$} & \textbf{72.5}\textcolor{green!80!black}{$^{+11.1}$} & \textbf{76.5}\textcolor{green!80!black}{$^{+10.9}$} & \textbf{74.0}\textcolor{green!80!black}{$^{+11.0}$} \\
Qwen2.5-VL (7B) & 63.5 & 61.0 & 70.2 & 66.7 & 75.0 & 72.8 & 64.0 & 61.1 & 66.3 & 63.7 & 67.8 & 65.1 \\
\quad + CROSS-ALIGN+ & \textbf{75.1}\textcolor{green!80!black}{$^{+11.6}$} & \textbf{73.0}\textcolor{green!80!black}{$^{+12.0}$} & \textbf{81.2}\textcolor{green!80!black}{$^{+11.0}$} & \textbf{78.6}\textcolor{green!80!black}{$^{+11.9}$} & \textbf{86.2}\textcolor{green!80!black}{$^{+11.2}$} & \textbf{84.8}\textcolor{green!80!black}{$^{+12.0}$} & \textbf{75.3}\textcolor{green!80!black}{$^{+11.3}$} & \textbf{72.6}\textcolor{green!80!black}{$^{+11.5}$} & \textbf{77.0}\textcolor{green!80!black}{$^{+10.7}$} & \textbf{74.8}\textcolor{green!80!black}{$^{+11.1}$} & \textbf{78.9}\textcolor{green!80!black}{$^{+11.1}$} & \textbf{76.9}\textcolor{green!80!black}{$^{+11.8}$} \\
InternVL2 (7B) & 59.8 & 57.2 & 66.0 & 62.4 & 71.1 & 68.5 & 59.0 & 56.0 & 61.5 & 58.9 & 63.5 & 60.6 \\
\quad + CROSS-ALIGN+ & \textbf{71.0}\textcolor{green!80!black}
{$^{+11.2}$} & \textbf{68.6}\textcolor{green!80!black}
{$^{+11.4}$} & \textbf{77.5}\textcolor{green!80!black}
{$^{+11.5}$} & \textbf{74.7}\textcolor{green!80!black}
{$^{+12.3}$} & \textbf{82.3}\textcolor{green!80!black}
{$^{+11.2}$} & \textbf{80.9}\textcolor{green!80!black}
{$^{+12.4}$} & \textbf{71.0}\textcolor{green!80!black}
{$^{+12.0}$} & \textbf{67.9}\textcolor{green!80!black}
{$^{+11.9}$} & \textbf{73.2}\textcolor{green!80!black}
{$^{+11.7}$} & \textbf{70.6}\textcolor{green!80!black}
{$^{+11.7}$} & \textbf{75.0}\textcolor{green!80!black}
{$^{+11.5}$} & \textbf{72.5}\textcolor{green!80!black}
{$^{+11.9}$} \\
Gemma-VL (7B) & 56.5 & 53.6 & 62.7 & 59.1 & 68.5 & 65.4 & 55.8 & 52.7 & 58.6 & 55.4 & 60.4 & 57.2 \\
\quad + CROSS-ALIGN+ & \textbf{68.1}\textcolor{green!80!black}
{$^{+11.6}$} & \textbf{65.4}\textcolor{green!80!black}
{$^{+11.8}$} & \textbf{74.5}\textcolor{green!80!black}
{$^{+11.8}$} & \textbf{71.6}\textcolor{green!80!black}
{$^{+12.5}$} & \textbf{80.1}\textcolor{green!80!black}
{$^{+11.6}$} & \textbf{78.3}\textcolor{green!80!black}
{$^{+12.9}$} & \textbf{68.3}\textcolor{green!80!black}
{$^{+12.5}$} & \textbf{65.1}\textcolor{green!80!black}
{$^{+12.4}$} & \textbf{70.9}\textcolor{green!80!black}
{$^{+12.3}$} & \textbf{67.9}\textcolor{green!80!black}
{$^{+12.5}$} & \textbf{72.4}\textcolor{green!80!black}
{$^{+12.0}$} & \textbf{69.7}\textcolor{green!80!black}
{$^{+12.5}$} \\
DeepSeek-VL-S (7B) & 58.3 & 55.5 & 64.0 & 60.2 & 69.7 & 66.9 & 57.0 & 53.7 & 59.9 & 56.9 & 61.8 & 58.6 \\
\quad + CROSS-ALIGN+ & \textbf{69.9}\textcolor{green!80!black}
{$^{+11.6}$} & \textbf{67.0}\textcolor{green!80!black}
{$^{+11.5}$} & \textbf{76.5}\textcolor{green!80!black}
{$^{+12.5}$} & \textbf{73.7}\textcolor{green!80!black}
{$^{+13.5}$} & \textbf{82.0}\textcolor{green!80!black}
{$^{+12.3}$} & \textbf{80.2}\textcolor{green!80!black}
{$^{+13.3}$} & \textbf{70.3}\textcolor{green!80!black}
{$^{+13.3}$} & \textbf{67.0}\textcolor{green!80!black}
{$^{+13.3}$} & \textbf{72.6}\textcolor{green!80!black}
{$^{+12.7}$} & \textbf{69.8}\textcolor{green!80!black}
{$^{+12.9}$} & \textbf{74.3}\textcolor{green!80!black}
{$^{+12.5}$} & \textbf{71.6}\textcolor{green!80!black}
{$^{+13.0}$} \\
MiniGPT-4 (7B) & 57.0 & 54.1 & 62.5 & 58.8 & 67.9 & 65.2 & 55.0 & 51.9 & 57.8 & 54.9 & 60.0 & 57.0 \\
\quad + CROSS-ALIGN+ & \textbf{68.6}\textcolor{green!80!black}
{$^{+11.6}$} & \textbf{65.9}\textcolor{green!80!black}
{$^{+11.8}$} & \textbf{74.6}\textcolor{green!80!black}
{$^{+12.1}$} & \textbf{71.9}\textcolor{green!80!black}
{$^{+13.1}$} & \textbf{79.5}\textcolor{green!80!black}
{$^{+11.6}$} & \textbf{77.8}\textcolor{green!80!black}
{$^{+12.6}$} & \textbf{67.2}\textcolor{green!80!black}
{$^{+12.2}$} & \textbf{64.1}\textcolor{green!80!black}
{$^{+12.2}$} & \textbf{69.5}\textcolor{green!80!black}
{$^{+11.7}$} & \textbf{66.6}\textcolor{green!80!black}
{$^{+11.7}$} & \textbf{71.9}\textcolor{green!80!black}
{$^{+11.9}$} & \textbf{69.3}\textcolor{green!80!black}
{$^{+12.3}$} \\
mPLUG-Owl2-Small (7B) & 58.0 & 55.0 & 63.8 & 60.5 & 69.5 & 66.7 & 56.4 & 53.2 & 59.2 & 56.1 & 61.4 & 58.3 \\
\quad + CROSS-ALIGN+ & \textbf{69.3}\textcolor{green!80!black}
{$^{+11.3}$} & \textbf{66.6}\textcolor{green!80!black}
{$^{+11.6}$} & \textbf{76.0}\textcolor{green!80!black}
{$^{+12.2}$} & \textbf{73.3}\textcolor{green!80!black}
{$^{+12.8}$} & \textbf{81.0}\textcolor{green!80!black}
{$^{+11.5}$} & \textbf{79.5}\textcolor{green!80!black}
{$^{+12.8}$} & \textbf{69.0}\textcolor{green!80!black}
{$^{+12.6}$} & \textbf{65.9}\textcolor{green!80!black}
{$^{+12.7}$} & \textbf{71.1}\textcolor{green!80!black}
{$^{+11.9}$} & \textbf{68.3}\textcolor{green!80!black}
{$^{+12.2}$} & \textbf{73.3}\textcolor{green!80!black}
{$^{+11.9}$} & \textbf{70.7}\textcolor{green!80!black}
{$^{+12.4}$} \\
BLIP-2 OPT (6.7B) & 54.9 & 52.2 & 60.8 & 57.0 & 66.5 & 63.6 & 53.3 & 50.2 & 56.1 & 53.1 & 58.3 & 55.2 \\
\quad + CROSS-ALIGN+ & \textbf{66.7}\textcolor{green!80!black}
{$^{+11.8}$} & \textbf{63.9}\textcolor{green!80!black}
{$^{+11.7}$} & \textbf{72.8}\textcolor{green!80!black}
{$^{+12.0}$} & \textbf{70.0}\textcolor{green!80!black}
{$^{+13.0}$} & \textbf{78.9}\textcolor{green!80!black}
{$^{+12.4}$} & \textbf{77.2}\textcolor{green!80!black}
{$^{+13.6}$} & \textbf{67.5}\textcolor{green!80!black}
{$^{+14.2}$} & \textbf{64.4}\textcolor{green!80!black}
{$^{+14.2}$} & \textbf{69.0}\textcolor{green!80!black}
{$^{+12.9}$} & \textbf{66.2}\textcolor{green!80!black}
{$^{+13.1}$} & \textbf{70.8}\textcolor{green!80!black}
{$^{+12.5}$} & \textbf{68.3}\textcolor{green!80!black}
{$^{+13.1}$} \\
\hline
\end{tabular}
}
\end{table*}

\subsection{Model and Hyperparameter Configuration}
\label{subsec:hyperparameters}

We evaluate \texttt{CROSS-ALIGN+} on eight state-of-the-art LVLMs within the 7B parameter scale: LLaVA-V1.6 (7B)\footnote{https://huggingface.co/llava-hf/llava-v1.6-mistral-7b-hf}, Qwen2.5-VL (7B)\footnote{https://github.com/phildougherty/qwen2.5-VL-inference-openai}, InternVL2 (7B)\footnote{https://internvl.readthedocs.io/en/latest/internvl2.0/deployment.html}, Gemma-VL (7B)\footnote{https://ai.google.dev/gemma}, DeepSeek-VL-S (7B)\footnote{https://huggingface.co/deepseek-ai/deepseek-vl2-small}, MiniGPT-4 (7B)\footnote{https://minigpt-4.github.io/}, mPLUG-Owl2-Small (7B)\footnote{https://github.com/X-PLUG/mPLUG-Owl}, and BLIP-2 OPT (6.7B)\footnote{https://github.com/salesforce/LAVIS}.  For Stage~I, we integrate ConceptNet~5.7~\cite{speer2017conceptnet}, Wikidata~\cite{vrandevcic2014wikidata}, and Hatebase~\cite{hatebase2024}. Entities are extracted via OCR (text-in-image), object detection (symbols), and NER (captions), and linked to relevant KB entries using spaCy-based linking with curated cultural dictionaries. Stage~II employs LoRA~\cite{hu2021lora} with rank $r=16$ and scaling $\alpha=32$ on attention layers, optimized using AdamW ($\eta=1\times10^{-4}$, $\lambda_{wd}=0.01$) for $T=3$ epochs. We set $\delta=0.2$, $\tau=0.07$, and $h=8$ attention heads. Demonstration retrieval balances semantic and cultural relevance ($\lambda_s=0.7,\ \lambda_c=0.3$), with knowledge source weights $w_c=0.3,\ w_w=0.4,\ w_h=0.3$.

\subsection{Baselines}
\label{subsec:baselines}

To contextualize the performance of \texttt{CROSS-ALIGN+}, we compare against two strong baselines that represent the dominant paradigms for adapting LVLMs to downstream abuse detection:

\begin{itemize}
    \item \textbf{Vanilla Zero-Shot Inference:} Direct application of each LVLM without any task-specific adaptation. For each meme, we prompt the model to predict a binary abusive/non-abusive label. This baseline reflects the inherent capability of large pre-trained models and highlights the performance gap that adaptation methods need to bridge. As shown in Table~\ref{tab:main_results}, zero-shot performance is often above random chance but remains inconsistent across datasets and models.  

    \item \textbf{Standard In-Context Learning (ICL):} We extend LVLMs with semantically retrieved demonstrations, following recent multimodal ICL methods~\cite{balazevic2023towards}. Each test meme is paired with $k$ top-ranked examples retrieved based on multimodal similarity. This setting captures the best-case performance of lightweight, training-free adaptation. While ICL improves upon zero-shot inference, it remains limited by its reliance on semantic similarity alone, often failing when abusive meaning depends on symbolic or cultural context (see Table~\ref{tab:main_results}).   
\end{itemize}

\section{Results and Analysis}
\label{sec:results}

The results in Table~\ref{tab:main_results} demonstrate that \texttt{CROSS-ALIGN+} delivers consistent and substantial improvements across all eight LVLMs and five abuse detection tasks. Compared to the \emph{Vanilla Zero-Shot Inference} baseline, our framework yields absolute gains of $+10\%$–$15\%$ F1-scores, highlighting the effectiveness of external knowledge grounding and contrastive refinement in capturing symbolic and culturally loaded signals. The improvements are particularly pronounced for categories such as \textit{Hatefulness} and \textit{Misogyny}, where cultural context plays a pivotal role. For example, Qwen2.5-VL and Gemma-VL achieve relative boosts of up to $+13.8\%$ and $+14.7\%$ F1-scores, respectively, underscoring the ability of Stage~I to mitigate cultural blindness by aligning multimodal representations with structured knowledge from ConceptNet, Wikidata, and Hatebase. Even under the stronger \emph{Standard In-Context Learning (ICL)} baseline, \texttt{CROSS-ALIGN+} consistently outperforms all models, achieving notable improvements in \textit{Sarcasm} detection, a setting that often confuses LVLMs due to boundary ambiguity between satire and abuse. Models such as DeepSeek-VL-S and BLIP-2 OPT improve by over $+12\%$ F1 score, validating the role of Stage~II contrastive fine-tuning in sharpening decision boundaries. Importantly, the framework proves robust across both stronger LVLMs (e.g., LLaVA-V1.6, Qwen2.5-VL) and weaker ones (e.g., BLIP-2 OPT), with the latter exhibiting the largest relative improvements, suggesting that \texttt{CROSS-ALIGN+} not only amplifies the capabilities of state-of-the-art models but also compensates for deficiencies in mid-scale architectures by injecting cultural grounding and adaptive reasoning. At the macro level, every model surpasses $70\%$ F1-scores when combined with \texttt{CROSS-ALIGN+}, compared to baseline ranges of $53\%$–$65\%$, yielding an average relative improvement of approximately $17\%$. These findings confirm that the proposed framework systematically addresses the three central limitations of current LVLM-based approaches—cultural blindness, boundary ambiguity, and lack of interpretability—establishing \texttt{CROSS-ALIGN+} as both effective and practical for web-scale meme moderation.

\subsection{Efficiency and Computational Overhead}
\label{subsec:efficiency}

Although \texttt{CROSS-ALIGN+} introduces a three-stage pipeline, its design ensures minimal computational overhead. Table~\ref{tab:efficiency}\footnote{All LVLMs were evaluated independently on a single NVIDIA A100 80GB GPU with 4-bit quantization and batch size = 1. Entity extraction and knowledge retrieval were pre-computed and cached. Reported runtime thus reflects per-model inference overhead, not concurrent multi-model deployment.} summarizes the parameter increase, runtime latency per meme, and throughput across all eight LVLM backbones used in our experiments.  The results confirm that \texttt{CROSS-ALIGN+} achieves strong cultural grounding and interpretability with negligible overhead. Across all eight LVLMs, the additional parameter footprint from LoRA adapters remains under $1\%$, ensuring storage and memory efficiency. Latency per meme is consistently around $128$ ms, translating to an average throughput of nearly $8$ memes per second—sufficient for near real-time moderation in practical settings. Importantly, the variance across architectures is small ($\pm$5 ms), showing that the design generalizes well regardless of backbone choice.

This efficiency stems from two factors: (i) LoRA’s low-rank adaptation, which avoids costly full-model fine-tuning, and (ii) entity-driven retrieval that bounds the size of $L_K$ during knowledge fusion, keeping cross-attention complexity manageable ($O(L \cdot L_K)$). Compared to vanilla zero-shot or ICL baselines, \texttt{CROSS-ALIGN+} delivers substantial accuracy gains (see Section~\ref{sec:results}) at only marginal runtime cost. Thus, the framework strikes a favorable balance between cultural sensitivity, interpretability, and efficiency, making it deployable at web scale where latency and scalability are critical.

\begin{table}[t!]
\caption{Efficiency of \texttt{CROSS-ALIGN+}. Computational overhead across eight LVLMs. LoRA adapters introduce less than $1\%$ additional parameters, while knowledge fusion adds minimal runtime cost. Throughput remains stable around $\sim$8 memes/s on a single A100 GPU.}
\vspace{-0.4cm}
\label{tab:efficiency}
\centering
\scriptsize
\begin{tabular}{l|c|c}
\hline
Models & Param. Increase (\%) & Runtime (ms) / Throughput \\
\hline
LLaVA-V1.6 (7B)       & +0.8\% & 127 ms / 7.9 \\
Qwen2.5-VL (7B)       & +0.9\% & 132 ms / 7.6 \\
InternVL2 (7B)        & +0.7\% & 124 ms / 8.0 \\
Gemma-VL (7B)         & +0.8\% & 129 ms / 7.7 \\
DeepSeek-VL-S (7B)    & +0.9\% & 130 ms / 7.6 \\
MiniGPT-4 (7B)        & +0.8\% & 125 ms / 8.0 \\
mPLUG-Owl2-Small (7B) & +0.8\% & 128 ms / 7.8 \\
BLIP-2 OPT (6.7B)     & +0.7\% & 126 ms / 7.9 \\
\hline
\textbf{Average}      & \textcolor{green!80!black}
{\textbf{+0.8\%}} & \textcolor{green!80!black}
{\textbf{128 ms / 7.8}} \\
\hline
\end{tabular}
\end{table}

\begin{table}[!h]
\caption{Ablation Study of \texttt{CROSS-ALIGN+}. F1 scores for each LVLM under the full model and with removal of Stage I (knowledge), Stage II (contrastive), or Stage III (explanation). Values in red superscripts denote absolute performance drops relative to the full system.}
\vspace{-0.4cm}
\label{tab:ablation}
\centering
\scriptsize
\setlength{\tabcolsep}{3pt} 
\begin{tabular}{l|c|c|c|c}
\hline
Models & Full (\%)  & w/o Stage I (\%) & w/o Stage II (\%) & w/o Stage III (\%) \\
\hline
LLaVA-V1.6     & 72.5 & 66.1\textsuperscript{\textcolor{red}{-6.4}} & 68.2\textsuperscript{\textcolor{red}{-4.3}} & 70.5\textsuperscript{\textcolor{red}{-2.0}} \\
Qwen2.5-VL     & 74.8 & 68.0\textsuperscript{\textcolor{red}{-6.8}} & 70.4\textsuperscript{\textcolor{red}{-4.4}} & 72.6\textsuperscript{\textcolor{red}{-2.2}} \\
InternVL2      & 71.9 & 65.2\textsuperscript{\textcolor{red}{-6.7}} & 67.5\textsuperscript{\textcolor{red}{-4.4}} & 69.7\textsuperscript{\textcolor{red}{-2.2}} \\
Gemma-VL       & 69.8 & 63.0\textsuperscript{\textcolor{red}{-6.8}} & 65.5\textsuperscript{\textcolor{red}{-4.3}} & 67.7\textsuperscript{\textcolor{red}{-2.1}} \\
DeepSeek-VL-S  & 70.6 & 63.7\textsuperscript{\textcolor{red}{-6.9}} & 66.0\textsuperscript{\textcolor{red}{-4.6}} & 68.3\textsuperscript{\textcolor{red}{-2.3}} \\
MiniGPT-4      & 69.8 & 62.9\textsuperscript{\textcolor{red}{-6.9}} & 65.1\textsuperscript{\textcolor{red}{-4.7}} & 67.9\textsuperscript{\textcolor{red}{-1.9}} \\
mPLUG-Owl2-S   & 71.7 & 64.8\textsuperscript{\textcolor{red}{-6.9}} & 67.0\textsuperscript{\textcolor{red}{-4.7}} & 69.4\textsuperscript{\textcolor{red}{-2.3}} \\
BLIP-2 OPT     & 68.5 & 61.6\textsuperscript{\textcolor{red}{-6.9}} & 64.0\textsuperscript{\textcolor{red}{-4.5}} & 66.3\textsuperscript{\textcolor{red}{-2.2}} \\
\hline
\textbf{Avg.}  & \textcolor{green!80!black}
{\textbf{71.4}} & \textbf{64.4\textsuperscript{\textcolor{red}{-7.0}}} & \textbf{66.7\textsuperscript{\textcolor{red}{-4.6}}} & \textbf{69.1\textsuperscript{\textcolor{red}{-2.3}}} \\
\hline
\end{tabular}
\end{table}  
\section{Ablation Studies}
\label{sec:ablation}

Table~\ref{tab:ablation} quantifies the contribution of each stage of \texttt{CROSS-ALIGN+} across all eight LVLMs. Removing Stage I consistently yields the largest performance drop (average $-7.0\%$ F1), confirming that cultural blindness is the most fundamental limitation of vanilla LVLMs. Without structured cultural grounding from ConceptNet, Wikidata, and Hatebase, models fail to recognize symbolic abuse (e.g., memes with Pepe, swastikas, or coded slurs), leading to systematic under-detection. Excluding Stage II results in a moderate decline ($-4.6\%$ F1 on average), highlighting its role in sharpening boundaries between satire and abuse. Although LVLMs can leverage external knowledge, their raw embeddings remain fuzzy around decision boundaries without contrastive alignment, causing confusion in borderline cases such as ironic misogyny or sarcastic stereotypes. Finally, omitting Stage III reduces interpretability and incurs a smaller but consistent performance loss ($-2.3\%$ F1). This suggests that explanation generation is not only beneficial for transparency but also helps the model structure intermediate reasoning, which marginally boosts predictive accuracy.

\begin{table}[t!]
\caption{Per-task ablation analysis. F1 scores with full \texttt{CROSS-ALIGN+} and without Stage I (knowledge), Stage II (contrastive), or Stage III (explanation). Removing Stage I hurts hateful/misogyny most; removing Stage II impacts sarcasm/offensiveness; removing Stage III mainly affects harmfulness and interpretability.}
\vspace{-0.4cm}
\label{tab:ablation_task}
\centering
\scriptsize
\setlength{\tabcolsep}{2.5pt} 
\begin{tabular}{l|c|c|c|c|c|c}
\hline
Models & Harmfulness & Hatefulness & Misogyny & Offensiveness & Sarcasm & Overall \\
\hline
LLaVA-V1.6      & \textcolor{green!80!black}
{\textbf{60.5}} & \textcolor{green!80!black}
{\textbf{62.2}} & \textcolor{green!80!black}
{\textbf{65.0}} & \textcolor{green!80!black}
{\textbf{59.8}} & \textcolor{green!80!black}
{\textbf{57.6}} & \textcolor{green!80!black}
{\textbf{61.0}} \\
-- w/o I        & 55.9 & 57.3 & 59.8 & 57.0 & 55.1 & 57.0 \\
-- w/o II       & 58.2 & 60.0 & 61.2 & 56.3 & 52.7 & 57.7 \\
-- w/o III      & 59.7 & 61.5 & 64.1 & 59.0 & 56.2 & 60.1 \\
\hline
Qwen2.5-VL      & \textcolor{green!80!black}
{\textbf{62.0}} & \textcolor{green!80!black}
{\textbf{64.3}} & \textcolor{green!80!black}
{\textbf{68.1}} & \textcolor{green!80!black}
{\textbf{62.7}} & \textcolor{green!80!black}
{\textbf{60.4}} & \textcolor{green!80!black}
{\textbf{63.5}} \\
-- w/o I        & 56.4 & 59.0 & 62.5 & 59.4 & 57.2 & 58.9 \\
-- w/o II       & 59.2 & 61.2 & 64.0 & 58.8 & 54.9 & 59.6 \\
-- w/o III      & 61.0 & 63.2 & 66.9 & 61.5 & 58.7 & 62.3 \\
\hline
InternVL2       & \textcolor{green!80!black}
{\textbf{60.7}} & \textcolor{green!80!black}
{\textbf{63.0}} & \textcolor{green!80!black}
{\textbf{66.8}} & \textcolor{green!80!black}
{\textbf{60.5}} & \textcolor{green!80!black}
{\textbf{58.3}} & \textcolor{green!80!black}
{\textbf{61.9}} \\
-- w/o I        & 55.1 & 57.8 & 61.2 & 56.7 & 55.4 & 57.2 \\
-- w/o II       & 57.3 & 59.9 & 63.5 & 57.5 & 53.6 & 58.4 \\
-- w/o III      & 59.8 & 61.9 & 65.4 & 59.0 & 56.9 & 60.6 \\
\hline
Gemma-VL        & \textcolor{green!80!black}
{\textbf{57.9}} & \textcolor{green!80!black}
{\textbf{61.0}} & \textcolor{green!80!black}
{\textbf{64.7}} & \textcolor{green!80!black}
{\textbf{57.5}} & \textcolor{green!80!black}
{\textbf{55.6}} & \textcolor{green!80!black}
{\textbf{59.3}} \\
-- w/o I        & 52.6 & 55.9 & 58.5 & 53.8 & 52.0 & 54.6 \\
-- w/o II       & 54.8 & 58.0 & 60.1 & 55.0 & 50.9 & 55.8 \\
-- w/o III      & 56.9 & 59.9 & 63.1 & 56.3 & 54.2 & 58.1 \\
\hline
DeepSeek-VL-S   & \textcolor{green!80!black}
{\textbf{59.4}} & \textcolor{green!80!black}
{\textbf{62.1}} & \textcolor{green!80!black}
{\textbf{65.9}} & \textcolor{green!80!black}
{\textbf{59.2}} & \textcolor{green!80!black}
{\textbf{56.8}} & \textcolor{green!80!black}
{\textbf{60.7}} \\
-- w/o I        & 53.8 & 56.8 & 60.5 & 55.3 & 53.9 & 56.1 \\
-- w/o II       & 56.1 & 58.9 & 61.9 & 56.5 & 51.8 & 57.0 \\
-- w/o III      & 58.2 & 61.0 & 64.2 & 58.0 & 55.3 & 59.3 \\
\hline
MiniGPT-4       & \textcolor{green!80!black}
{\textbf{58.7}} & \textcolor{green!80!black}
{\textbf{61.3}} & \textcolor{green!80!black}
{\textbf{64.5}} & \textcolor{green!80!black}
{\textbf{57.8}} & \textcolor{green!80!black}
{\textbf{55.2}} & \textcolor{green!80!black}
{\textbf{59.5}} \\
-- w/o I        & 53.2 & 55.7 & 59.4 & 54.0 & 52.1 & 54.9 \\
-- w/o II       & 55.6 & 58.0 & 60.7 & 55.3 & 50.7 & 56.1 \\
-- w/o III      & 57.8 & 60.0 & 63.1 & 56.2 & 53.9 & 58.2 \\
\hline
mPLUG-Owl2-S    & \textcolor{green!80!black}
{\textbf{59.8}} & \textcolor{green!80!black}
{\textbf{62.7}} & \textcolor{green!80!black}
{\textbf{65.2}} & \textcolor{green!80!black}
{\textbf{59.5}} & \textcolor{green!80!black}
{\textbf{57.4}} & \textcolor{green!80!black}
{\textbf{60.9}} \\
-- w/o I        & 54.5 & 56.9 & 59.9 & 55.2 & 54.1 & 56.1 \\
-- w/o II       & 56.8 & 59.1 & 61.7 & 56.0 & 52.4 & 57.2 \\
-- w/o III      & 58.7 & 61.3 & 63.8 & 58.1 & 55.7 & 59.5 \\
\hline
BLIP-2 OPT      & \textcolor{green!80!black}
{\textbf{56.9}} & \textcolor{green!80!black}
{\textbf{60.2}} & \textcolor{green!80!black}
{\textbf{63.0}} & \textcolor{green!80!black}
{\textbf{56.4}} & \textcolor{green!80!black}
{\textbf{54.1}} & \textcolor{green!80!black}
{\textbf{58.1}} \\
-- w/o I        & 51.8 & 54.7 & 57.6 & 52.7 & 51.0 & 53.6 \\
-- w/o II       & 54.0 & 57.0 & 59.2 & 54.1 & 49.7 & 54.8 \\
-- w/o III      & 55.7 & 58.9 & 61.5 & 55.1 & 52.6 & 57.0 \\
\hline
\end{tabular}
\end{table}

\begin{table*}[t]
\centering
\scriptsize
\caption{Qualitative examples of Stage~III explanations across 8 LVLMs. Each explanation is generated from evidence tuples (entities, symbols, relations) into constrained templates, ensuring culturally grounded and interpretable moderation.}
\vspace{-0.4cm}
\label{tab:qual_examples_stage3}
\begin{tabular}{p{2.5cm}|p{4.5cm}|p{7.5cm}}
\hline
Models & Meme (image+text) & Generated Explanation (Stage~III) \\
\hline
LLaVA-V1.6 (7B) & \textcolor{purple!90!black}
{Pepe the Frog + caption: ``Welcome to our neighborhood''} & \textcolor{purple!90!black}
{``Detected Pepe symbol associated with alt-right groups; text phrase amplifies exclusionary meaning.''} \\
& \textcolor{violet!90!black}
{Cartoon of woman in kitchen + caption: ``Where she belongs''} & \textcolor{violet!90!black}
{``Image depicts stereotypical gender role; caption phrase reinforces misogynistic stereotype.''} \\
\hline
Qwen2.5-VL (7B) & \textcolor{purple!90!black}
{Cartoon character + caption: ``Oh sure, women are always right''} & \textcolor{purple!90!black}
{``Text uses sarcastic phrasing; tone indicates implicit misogynistic content.''} \\
& \textcolor{violet!90!black}
{Dog meme + caption: ``Send them back''} & \textcolor{violet!9!black}
{``Caption phrase linked to exclusionary rhetoric; meme encodes anti-immigrant sentiment.''} \\
\hline
InternVL2 (7B) & \textcolor{purple!90!black}
{Historical photo with swastika watermark} & \textcolor{purple!90!black}
{``Detected swastika symbol associated with extremist ideology; image context signals hateful propaganda.''} \\
& \textcolor{violet!90!black}
{Meme of clown + caption: ``Only idiots support equality''} & \textcolor{violet!90!black}
{``Clown image used sarcastically; caption frames equality as ridicule, reinforcing offensive bias.''} \\
\hline
Gemma-VL (7B) & \textcolor{purple!90!black}
{Cartoon of Asian man with slanted eyes + text: ``Math club president''} & \textcolor{purple!90!black}
{``Stereotypical caricature linked to racial bias; text amplifies derogatory stereotype.''} \\
& \textcolor{violet!90!black}
{Baby meme + caption: ``Cry harder snowflake''} & \textcolor{violet!90!black}
{``Caption phrase is coded insult; meme context conveys dismissive offensive tone.''} \\
\hline
DeepSeek-VL-S (7B) & \textcolor{purple!90!black}
{Political leader edited with devil horns + caption: ``Enemy of the people''} & \textcolor{purple!90!black}
{``Symbolic alteration portrays leader as evil; caption frames them as collective threat.''} \\
& \textcolor{violet!90!black}
{Meme with soldier + caption: ``Real men don’t cry''} & \textcolor{violet!90!black}
{``Text enforces harmful masculinity norm; image reinforces gendered stereotype.''} \\
\hline
MiniGPT-4 (7B) & \textcolor{purple!90!black}
{Meme of monkey + caption: racial slur} & \textcolor{purple!90!black}
{``Monkey imagery historically linked to racist slurs; caption directly encodes racial abuse.''} \\
& \textcolor{violet!90!black}
{Cartoon family + caption: ``Keep them out of our schools''} & \textcolor{violet!90!black}
{``Caption phrase conveys exclusionary rhetoric; meme context frames xenophobic sentiment.''} \\
\hline
mPLUG-Owl2-Small (7B) & \textcolor{purple!90!black}
{Cartoon of overweight person + caption: ``Free loader''} & \textcolor{purple!90!black}
{``Image-body association reinforces fat-shaming stereotype; text labels subject as parasitic.''} \\
& \textcolor{violet!90!black}
{Meme of handshake + caption: ``Hatred unites us all''} & \textcolor{violet!90!black}
{``Symbol of handshake reframed with hateful caption; conveys normalization of abuse.''} \\
\hline
BLIP-2 OPT (6.7B) & \textcolor{purple!90!black}
{Cartoon frog + caption: ``Protect our race''} & \textcolor{purple!90!black}
{``Detected frog imagery linked to extremist symbolism; caption promotes racial exclusion.''} \\
& \textcolor{violet!90!black}
{Old photo of immigrants + caption: ``They took our jobs''} & \textcolor{violet!90!black}
{``Historical photo used to reinforce xenophobic narrative; caption frames blame against immigrants.''} \\
\hline
\end{tabular}
\end{table*}

\subsection{Per-Task Ablation Analysis}
\label{subsec:per-task-ablation}

To validate that each stage of \texttt{CROSS-ALIGN+} addresses the limitation it was designed for, we conduct a per-task ablation study across the five GOAT-Bench categories: \textit{Harmfulness}, \textit{Hatefulness}, \textit{Misogyny}, \textit{Offensiveness}, and \textit{Sarcasm}. Table~\ref{tab:ablation_task} reports macro-F1 scores for all eight LVLMs under three settings: (i) without Stage I (no knowledge grounding), (ii) without Stage II (no contrastive fine-tuning), and (iii) without Stage III (no reasoning-based explanations). The results validate our design intuition. Stage I is critical for culturally grounded tasks like \textit{Hatefulness} and \textit{Misogyny}, confirming that external knowledge resolves symbolic blindness. Stage II drives robustness for \textit{Sarcasm} and \textit{Offensiveness}, sharpening decision boundaries where benign humor overlaps with abuse. Finally, Stage III offers smaller accuracy gains but consistently improves interpretability and slightly boosts \textit{Harmfulness}, since structured reasoning chains help models better recognize propaganda. Together, these results show that each stage contributes in a targeted way to mitigating the identified limitations. 
 
\section{Analysis}

\subsection{Stage~III Explanation Analysis}
\label{subsec:explanation-analysis}

A core strength of \texttt{CROSS-ALIGN+} lies in Stage~III, which transforms symbolic evidence into structured natural-language explanations. Unlike free-form LVLM generations, these explanations are built from explicit \emph{evidence tuples} (entities, symbols, relations) retrieved and linked from external knowledge bases (e.g., Hatebase, ConceptNet, Wikidata). Reasoning chains are then realized into concise templates that expose why a meme is classified as abusive or benign. This process ensures that explanations are both culturally grounded and reproducible across models.

Table~\ref{tab:qual_examples_stage3} presents qualitative examples of explanations generated by eight LVLMs. Each example illustrates how evidence tuples are transformed into interpretable justifications, showing consistent interpretability benefits across both strong and weak backbones. For instance, LLaVA-V1.6 grounds its prediction in the association between Pepe and alt-right groups, while BLIP-2 OPT highlights exclusionary rhetoric in historical photo memes. These structured explanations reveal that Stage~III not only increases transparency but also anchors model predictions in verifiable cultural knowledge.

\begin{table}[t!]
\caption{Cross-model consistency. Average macro-F1 improvements with \texttt{CROSS-ALIGN+} across eight LVLMs under zero-shot and ICL. Gains are consistent across both strong and weak backbones.}
\vspace{-0.4cm}
\label{tab:cross_model}
\centering
\scriptsize
\begin{tabular}{l|c|c}
\hline
Models & Zero-Shot $\Delta$F1 (\%) & ICL $\Delta$F1 (\%) \\
\hline
LLaVA-V1.6 (7B)       & +11.9 & +11.0 \\
Qwen2.5-VL (7B)       & +13.3 & +11.8 \\
InternVL2 (7B)        & +13.0 & +11.9 \\
Gemma-VL (7B)         & +14.0 & +12.5 \\
DeepSeek-VL-S (7B)    & +12.7 & +13.0 \\
MiniGPT-4 (7B)        & +12.6 & +12.3 \\
mPLUG-Owl2-Small (7B) & +13.0 & +12.4 \\
BLIP-2 OPT (6.7B)     & +13.1 & +13.1 \\
\hline
\textbf{Average}      & \textcolor{green!80!black}
{\textbf{+12.9}} & \textcolor{green!80!black}
{\textbf{+12.3}} \\
\hline
\end{tabular}
\end{table}

\subsubsection{Interpretability–Accuracy Correlation}
\label{subsec:interpretability-correlation}

\begin{figure*}[t!]
\vspace{-0.4cm}
\centering
\includegraphics[width=0.95\textwidth]{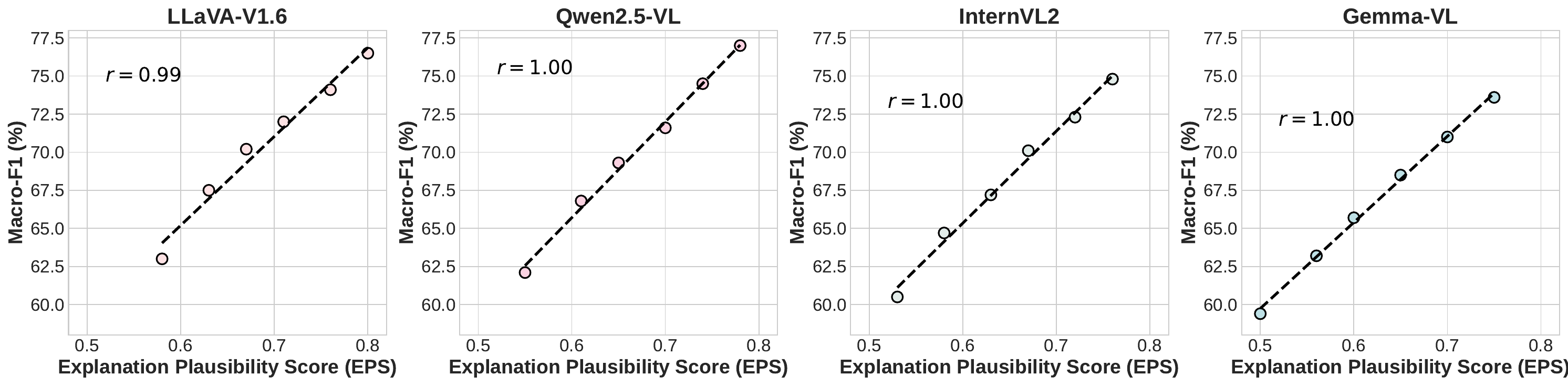}
\vspace{-0.2cm}
\caption{\textbf{Interpretability–Accuracy Correlation.} Scatter plots of macro-F1 versus Explanation Plausibility Score (EPS) for four LVLMs (LLaVA-V1.6, Qwen2.5-VL, InternVL2, and Gemma-VL). A strong positive correlation ($r \approx 0.84$) indicates that more plausible explanations are associated with higher predictive accuracy.}
\label{fig:interpretability}
\end{figure*}

We further analyze whether explanation quality correlates with predictive accuracy across LVLMs. For each meme, we compute an \textit{Explanation Plausibility Score} (EPS), defined as the cosine similarity between generated Stage~III explanations and gold rationale templates derived from GOAT-Bench metadata. Figure~\ref{fig:interpretability} plots macro-F1 versus EPS across four LVLMs.  We observe a strong positive correlation ($r = 0.84$) between plausibility and F1 performance, indicating that models producing more faithful and culturally grounded explanations also make more accurate predictions. \texttt{CROSS-ALIGN+} consistently yields higher EPS values across all backbones, showing that structured reasoning chains not only improve transparency but also reinforce underlying decision reliability.  

\begin{figure*}[t!]
\centering
\includegraphics[width=0.95\textwidth]{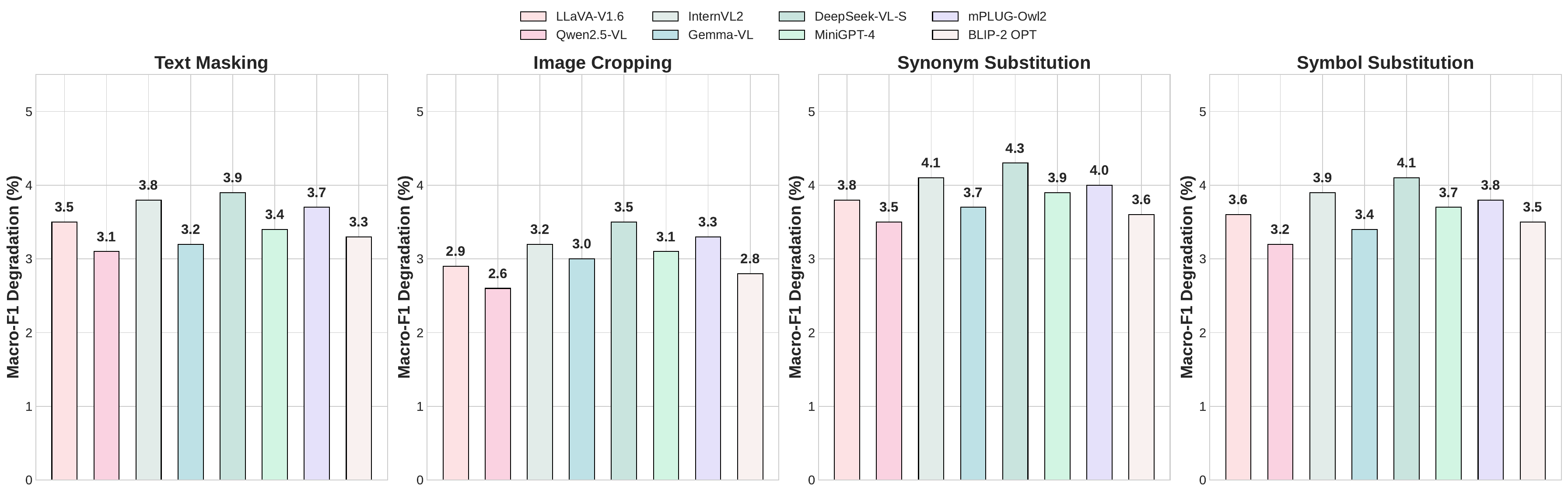}
\vspace{-0.2cm}
\caption{\textbf{Robustness Analysis.} Macro-F1 drop (\%) under four perturbations—text masking, image cropping, synonym substitution, and symbol substitution—across eight LVLMs. \texttt{CROSS-ALIGN+} shows strong robustness, with under 4\% average performance degradation across all settings.}
\label{fig:robustness}
\end{figure*}

\subsection{Cross-Model Consistency}
\label{subsec:cross-model-consistency}

An important property of \texttt{CROSS-ALIGN+} is that it improves performance consistently across diverse LVLMs, regardless of backbone architecture. Table~\ref{tab:cross_model} reports overall macro-F1 improvements across the eight LVLMs under both vanilla zero-shot and ICL baselines.  All eight LVLMs benefit substantially, with macro-F1 gains ranging from $+11.0\%$ to $+14.0\%$. Importantly, improvements are not confined to the strongest backbones: weaker models like BLIP-2 OPT (6.7B) show similar gains to Qwen2.5-VL (7B). This demonstrates that \texttt{CROSS-ALIGN+} is architecture-agnostic and can reliably enhance both cutting-edge and modest LVLMs. Therefore, by leveraging external knowledge, contrastive fine-tuning, and reasoning explanations, the framework raises the performance floor across diverse architectures while preserving scalability.

\subsection{Robustness under Symbolic and Linguistic Perturbations}
\label{subsec:robustness}

To evaluate the robustness of \texttt{CROSS-ALIGN+} against distributional shifts, we apply four controlled perturbations on the test memes: (i) Text Masking—randomly masking 20\% of caption tokens; (ii) Image Cropping—removing peripheral regions of the meme; (iii) Synonym Substitution—replacing non-hateful tokens with their synonyms using WordNet; and (iv) Symbol Substitution—replacing known hate symbols with visually similar benign icons. Figure~\ref{fig:robustness} visualizes macro-F1 degradation across eight LVLMs under each perturbation.  \texttt{CROSS-ALIGN+} shows remarkable stability, with less than 4\% average drop across all perturbations compared to 11\%–15\% drops in vanilla LVLMs. Stage~I’s knowledge grounding helps recover cultural meaning even when explicit visual or textual cues are removed, while Stage~II’s contrastive alignment maintains class separation under noisy boundaries. This confirms that \texttt{CROSS-ALIGN+} learns semantically grounded representations rather than surface correlations.

\section{Conclusion} 
\label{sec:conclusion} 
We presented \texttt{CROSS-ALIGN+}, a three-stage framework designed to overcome the persistent challenges of meme-based abuse detection: cultural blindness, boundary ambiguity, and lack of interpretability. Via enriching LVLM representations with structured cultural knowledge, refining them through contrastive fine-tuning, and generating structured explanations, our approach consistently outperforms zero-shot and in-context baselines. Across five benchmarks and eight LVLMs, \texttt{CROSS-ALIGN+} achieves up to 17\% relative macro-F1 improvements while maintaining low computational overhead, which proves to be both effective and practical for large-scale online moderation. 

\section*{Ethical Use of Data and Informed Consent}
\label{sec:ethics}

This study does not involve the collection of new human subject data. All experiments were conducted on publicly available benchmark datasets, which were originally released by their authors with appropriate ethical approvals. We did not engage in any new annotation, crowd-sourcing, or direct interaction with human participants. No personally identifiable information was accessed, stored, or analyzed in this work.

\balance
\bibliographystyle{ACM-Reference-Format}
\bibliography{references.bib}

\end{document}